\documentclass[10pt,twocolumn,letterpaper]{article}

\usepackage[accsupp]{axessibility}

\usepackage{iccv}
\usepackage{times}
\usepackage{epsfig}
\usepackage{graphicx}
\usepackage{amsmath}
\usepackage{amssymb}

\usepackage{setspace}

\usepackage{threeparttable}
\usepackage{adjustbox}
\usepackage{multirow}
\usepackage{mathtools}

\usepackage{booktabs}
\usepackage{subfigure}
\usepackage{bm}
\usepackage{makecell}
\usepackage{wrapfig}
\usepackage{algorithm}
\usepackage{algorithmic}
\usepackage{bbm}
\usepackage{epstopdf}
\usepackage{tablefootnote}
\usepackage[normalem]{ulem}
\usepackage{color}

\usepackage{algorithm}
\usepackage{algorithmic}

\usepackage[table]{xcolor}

\usepackage{bm}
\usepackage{mathrsfs}

\DeclareMathAlphabet{\mathcal}{OMS}{cmsy}{m}{n}

\usepackage[pagebackref=true,breaklinks=true,letterpaper=true,colorlinks,bookmarks=false]{hyperref}

\iccvfinalcopy 

\def\iccvPaperID{5647}

\ificcvfinal\fi

\begin{document}

\title{NAS-OoD: Neural Architecture Search for\\Out-of-Distribution Generalization}

\author{Haoyue Bai\\
The Hong Kong University\\
of Science and Technology\\
{\tt\small hbaiaa@cse.ust.hk}
\and
Fengwei Zhou\\
Huawei Noah's Ark Lab\\
{\tt\small zhoufengwei@huawei.com}
\and
Lanqing Hong\\
Huawei Noah's Ark Lab\\
{\tt\small honglanqing@huawei.com}
\and
Nanyang Ye \thanks{Nanyang Ye is the corresponding author.}\\
Shanghai Jiao Tong University\\
{\tt\small ynylincoln@sjtu.edu.cn}
\and
S.-H. Gary Chan\\
The Hong Kong University\\
of Science and Technology\\
{\tt\small gchan@cse.ust.hk}
\and
Zhenguo Li\\
Huawei Noah's Ark Lab\\
{\tt\small li.zhenguo@huawei.com}
}

\maketitle
\ificcvfinal\thispagestyle{empty}\fi

\begin{abstract}

Recent advances on Out-of-Distribution (OoD) generalization reveal the robustness of deep learning models against distribution shifts.
However, existing works focus on OoD algorithms, such as invariant risk minimization, domain generalization, or stable learning, without considering the influence of deep model architectures on OoD generalization, which may lead to sub-optimal performance. 
Neural Architecture Search (NAS) methods search for architecture based on its performance on the training data, which may result in poor generalization for OoD tasks.
In this work, we propose robust Neural Architecture Search for OoD generalization (NAS-OoD), which optimizes the architecture with respect to its performance on generated OoD data by gradient descent.
Specifically, a data generator is learned to synthesize OoD data by maximizing losses computed by different neural architectures, while the goal for architecture search is to find the optimal architecture parameters that minimize the synthetic OoD data losses. 
The data generator and the neural architecture are jointly optimized in an end-to-end manner, and the minimax training process effectively discovers robust architectures that generalize well for different distribution shifts.
Extensive experimental results show that NAS-OoD achieves superior performance on various OoD generalization benchmarks with deep models having a much fewer number of parameters. In addition, on a real industry dataset, the proposed NAS-OoD method reduces the error rate by more than $70\%$ compared with the state-of-the-art method, demonstrating the proposed method's practicality for real applications.

\end{abstract}


\section{Introduction} \label{sec:intro}

\begin{figure}
    \centering
    \includegraphics[width=0.73\linewidth]{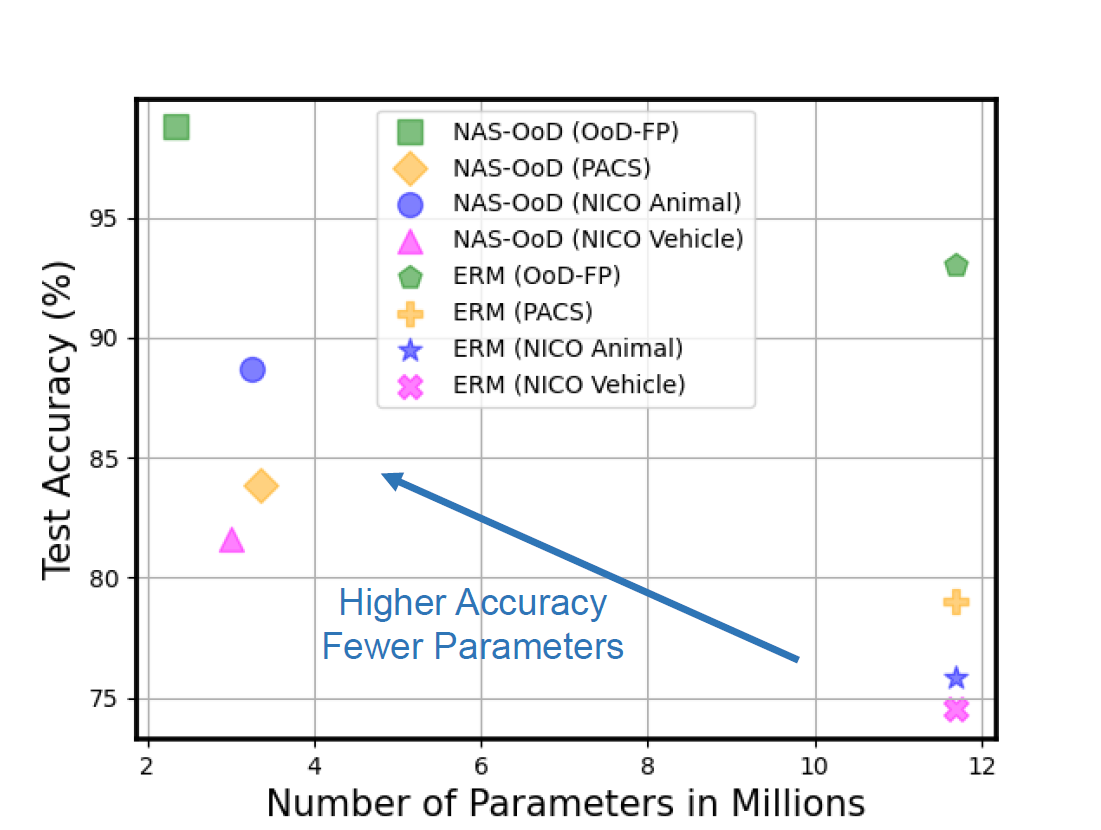}
    \caption{NAS-OoD performs significantly better than existing OoD generalization baselines in terms of test accuracy and network parameter numbers.
    The upper left points are better than lower right ones
    because they have higher test accuracy and lower parameter numbers.}
    \label{fig:testacc}
\end{figure}

Deep learning models have encountered significant performance drop in Out-of-Distribution (OoD) scenarios~\cite{bahng2019rebias,krueger2020outofdistribution},
where test data come from a distribution different from that of the training data. 
With their growing use in real-world applications in which mismatches of test and training data distributions are often observed~\cite{koh2020wilds}, extensive efforts have been devoted to improving generalization ability
~\cite{Li2017,arjovsky2019invariant,he2020towards,bai2020decaug}. Risk regularization methods~\cite{arjovsky2019invariant,ahuja2020invariant,xie2020risk}
aim to learn invariant representations across different training environments by imposing different invariant risk regularization. Domain generalization methods~\cite{Li2017,li2017learning,carlucci2019domain,zhou2020learning} learn models from multiple domains such that they can generalize well to unseen domains. Stable learning~\cite{Kuang2018,kuang2020stable,he2020towards} focuses on identifying stable and causal features for predictions.
Existing works, however, seldom consider the effects of architectures on generalization ability. 
On the other hand, some pioneer works suggest that different architectures show varying OoD generalization abilities~\cite{hendrycks2020pretrained, dapello2020simulating, li2020network}.
How a network's architecture affects its ability to handle OoD distribution shifts is still an open problem.

Conventional Neural Architecture Search (NAS) methods search for architectures with maximal predictive performance on the validation data that are randomly divided from the training data~\cite{zoph2017neural,pham2018efficient,liu2018darts,yang2020ista}. The discovered architectures are supposed to perform well on unseen test data under the assumption that data are Independent and Identically Distributed (IID).
While novel architectures discovered by recent NAS methods have demonstrated superior performance on different tasks with the IID assumption~\cite{tan2019efficientnet,ghiasi2019fpn,yao2019sm,hu2020count}, they may suffer from over-fitting in OoD scenarios, where the test data come from another distribution. 
A proper validation set that can evaluate the performance of architectures on the test data with distribution shifts is crucial in OoD scenarios.

\begin{figure*}
    \centering
    \includegraphics[width=0.88\linewidth]{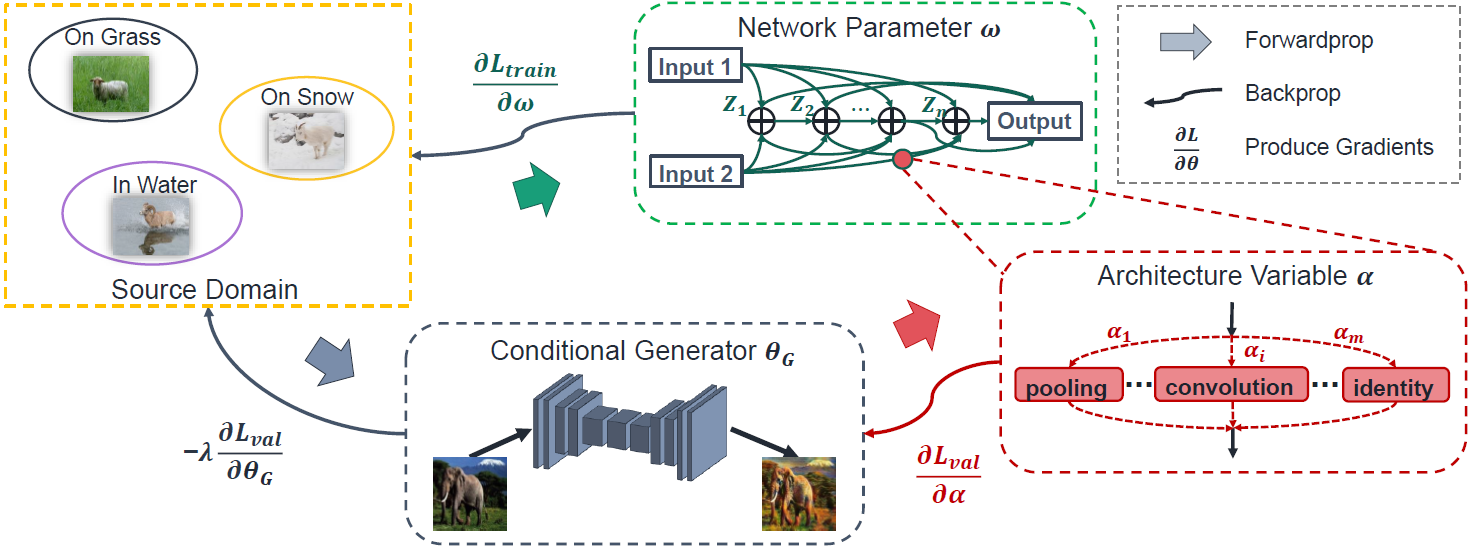}
    \caption{An overview of the proposed NAS-OoD. A conditional generator is learned to map the original training data to synthetic OoD examples by maximizing their losses computed by different neural architectures. Meanwhile, the architecture search process is optimized by minimizing the synthetic OoD data losses.
    }
    \label{fig:framework}
\end{figure*}

In this paper, we propose robust NAS for OoD generalization (NAS-OoD) that searches architectures with maximal predictive performance on OoD examples generated by a conditional generator. An overview of the proposed method is illustrated in Figure~\ref{fig:framework}.
To do NAS and train an OoD model simultaneously, we follow the line of gradient-based methods for NAS~\cite{liu2018darts,xie2018snas,cai2018proxylessnas,hu2020dsnas,yang2020ista}, however, we extend that on several fronts. The discrete selection of architectures is relaxed to be differentiable by building all candidate architectures into a supernet with parameter sharing and adopting a softmax choice over all possible network operations. The goal for architecture search is to find the optimal architecture parameters that minimize the validation loss under the condition that the corresponding network parameters minimize the training loss. 

Instead of using part of the training set as the validation set, we train a conditional generator to map the original training data to synthetic OoD examples as the validation data. The parameters of the generator are updated to maximize the validation loss computed by the supernet. This update encourages the generator to synthesize data having a different distribution from the original training data since the supernet is optimized to minimize the error on the training data. To search for the architectures with optimal OoD generalization ability, the architecture parameters are optimized to minimize the loss on the validation set containing synthetic OoD data. This minimax training process effectively drives both the generator and architecture search to improve their performance and finally derive the robust architectures that perform well for OoD generalization.

Our main contributions can be summarized as follows:
\begin{enumerate}
    \item To the best of our knowledge, NAS-OoD is the first attempt to introduce NAS for OoD generalization, where a conditional generator is jointly optimized to synthesize OoD examples helping to correct the supervisory signal for architecture search.
    \item NAS-OoD gets the optimal architecture and all optimized parameters in a single run. The minimax training process effectively discovers robust architectures that generalize well for different distribution shifts.
    \item We take the first step to understanding the OoD generalization of neural network architectures systematically. We provide a statistical analysis of the searched architectures and our preliminary practice shows that architecture does influence OoD robustness.
    \item Extensive experimental results show that NAS-OoD outperforms the previous SOTA methods 
    and achieves the best overall OoD generalization performance on various types of OoD tasks with the discovered architectures having a much fewer number of parameters.
\end{enumerate}

\section{Related Work} \label{sec:rela}

\subsection{Out-of-Distribution Generalization}
\label{rela:ood}
Data distribution mismatches between training and testing set exist in many real-world scenes. Different methods have been developed to tackle OoD shifts. IRM~\cite{arjovsky2019invariant} targets to extract invariant representation from different environments via an invariant risk regularization. IRM-Games~\cite{ahuja2020invariant} aims to achieve the Nash equilibrium among multiple environments to find invariants based on ensemble methods. REx~\cite{krueger2020outofdistribution} proposes a min-max procedure to deal with the worst linear combination of risks across different environments. MASF~\cite{dou2019domain} adopts a framework to learn invariant features among domains. JiGen~\cite{carlucci2019domain} jointly classifies objects and solves unsupervised jigsaw tasks. CuMix~\cite{mancini2020towards} aims to recognize unseen categories in unseen domains through a curriculum procedure to mix up data and labels from different domains. DecAug~\cite{bai2020decaug} proposes a decomposed feature representation and semantic augmentation approach to address diversity and distribution shifts jointly. The work of \cite{hendrycks2019using} finds that using pre-training can improve model robustness and uncertainty. However, existing OoD generalization approaches seldom consider the effects of architecture which leads to suboptimal performances. In this work, we propose NAS-OoD, a robust network architecture search method for OoD generalization.

\subsection{Neural Architecture Search}
EfficientNet~\cite{tan2019efficientnet} proposes a new scaling method that uniformly scales all dimensions of depth, width, and resolution via an effective compound coefficient. EfficientNet design a new baseline which achieves much better accuracy and efficiency than previous ConvNets. One-shot NAS~\cite{bender2018understanding} discusses the weight sharing scheme for one-shot architecture search and shows that it is possible to identify promising architectures without either hypernetworks or RL efficiently. DARTS~\cite{liu2018darts} presents a differentiable manner to deal with the scalability challenge of architecture search. ISTA-NAS~\cite{yang2020ista} formulates neural architecture search as a sparse coding problem. In this way, the network in search satisfies the sparsity constraint at each update and is efficient to train. SNAS~\cite{xie2018snas} reformulates NAS as an optimization problem on parameters of a joint distribution for the search space in a cell. DSNAS~\cite{hu2020dsnas} proposes an efficient NAS framework that simultaneously optimizes architecture and parameters with a low-biased Monte Carlo estimate. NASDA~\cite{li2020network} leverages a principle framework that uses differentiable neural architecture search to derive optimal network architecture for domain adaptation tasks.
NADS~\cite{ardywibowo2020nads} learns a posterior distribution on the architecture search space to enable uncertainty quantification for better OoD detection and aims to spot anomalous samples.
The work \cite{chen2020robustness} uses a robust loss to mitigate the performance degradation under symmetric label noise. However, 
NAS overfits easily, the work \cite{yang2019evaluation, guo2020single} points out that NAS evaluation is frustratingly hard. Thus, it is highly non-trivial to extend existing NAS algorithms to the OoD setting.

\subsection{Robustness from Architecture Perspective}
\label{rela:arch}
Recent studies show that different architectures present different generalization abilities.
The work of \cite{zhang2021can} uses a functional modular probing method to analyze deep model structures under the OoD setting.
The work \cite{hendrycks2020pretrained} examines and shows that pre-trained transformers achieve not only high accuracy on in-distribution examples but also improvement of out-of-distribution robustness. The work \cite{dapello2020simulating} presents CNN models with neural hidden layers that better simulate the primary visual cortex improve robustness against image perturbations. The work \cite{dosovitskiy2020image} uses a pure transformer applied directly to sequences of image patches, which performs quite well on image classification tasks compared with relying on CNNs. The work of \cite{dong2020adversarially} targets to improve the adversarial robustness of the network with NAS and achieves superior performance under various attacks. However, they do not consider OoD generalization from the architecture perspective.

\section{Methodology} \label{sec:meth}

In this section, we first introduce preliminaries on conventional NAS and their limitations in OoD scenarios (Section~\ref{meth:nas}). Then, we describe the details of our robust Neural Architecture Search for OoD generalization (Section~\ref{meth:nas-ood}). 

\subsection{Preliminaries on Differentiable Neural \\Architecture Search}
\label{meth:nas}

Conventional NAS methods mainly search for computation cells as the building units to construct a network~\cite{liu2018darts, xie2018snas, hu2020dsnas}. The search space of a cell is defined as a directed acyclic graph with $n$ ordered nodes $\{z_1, z_2, ..., z_n\}$ and edges $\mathcal{\xi}=\{e^{i,j}|1 \le i < j \le n\}$. Each edge includes $m$ candidate network operations chosen from a pre-defined operation space $\mathcal{O} = \{o_1, o_2, ..., o_m\}$, such as max-pooling, identity and dilated convolution. The binary variable $s_{k}^{(i,j)} \in \{0, 1\}$ denotes the corresponding active connection. Thus, the node can be formed as:
\begin{equation}
z_j = \sum_{i=1}^{j-1}\sum_{k=1}^{m}s_k^{(i,j)}o_k(z_i) = \boldsymbol{s}_{j}^{T}\boldsymbol{o}_j,
\end{equation}
where $\boldsymbol{s}_{j}$ is the vector consists of $s_k^{(i,j)}$ and $\boldsymbol{o}_j$ denotes the vector formed by $o_k(z_i)$. As the binary architecture variables $s_j$ is hard to optimize in a differentiable way, recent DARTS-based NAS methods make use of the continuous relaxation in the form of
\begin{equation}
s_k^{(i,j)}=\text{exp}(\alpha_k^{(i,j)})/\sum_k\text{exp}(\alpha_k^{(i,j)}),
\end{equation}
and optimize $\alpha_k^{(i,j)}$ as trainable architecture parameters~\cite{liu2018darts}, which can be formulated as the following bilevel optimization problem:
\begin{equation}
\begin{aligned}
&\boldsymbol{\alpha}^* = \mathop{\arg \min}_{\boldsymbol{\alpha}} \ \ell_{\text{val}}(\boldsymbol{\omega}^{\ast}(\boldsymbol{\alpha}), \boldsymbol{\alpha}), \\
&\mathop{s.t.}\quad \boldsymbol{\omega}^{\ast}(\boldsymbol{\alpha}) = \mathop{\arg \min}_{\boldsymbol{\omega}}\ \ell_{\text{train}}(\boldsymbol{\omega}, \boldsymbol{\alpha}),
\end{aligned}
\end{equation}
where $\boldsymbol{\alpha}$ denotes the architecture variable vector formed by $\alpha_k^{(i,j)}$, and $\boldsymbol{\omega}$ denotes the supernet parameters. 
$\ell_{\text{train}}$ and $\ell_{\text{val}}$ denote the training and validation losses, respectively. In the search phase, $\boldsymbol{\alpha}$ and $\boldsymbol{\omega}$ are optimized in an alternate manner.

The validation data used for the above architecture search method are usually divided from the training data. Previous research demonstrates that the derived architectures perform well on different tasks~\cite{liu2018darts, xie2018snas, hu2020dsnas} when the training and test data are IID. 
However, when dealing with OoD tasks, where the test data come from another distribution, using part of the training set as the validation set may cause NAS methods suffer from over-fitting and the searched architectures to be sub-optimal in this situation. Thus, a proper validation set is needed to effectively evaluate the performance of discovered architectures on the test set in OoD scenarios.

\subsection{NAS-OoD: Neural Architecture Search for\\ OoD Generalization}
\label{meth:nas-ood}

In OoD learning tasks, we are provided with $K$ source domains. The target is to discover the optimal network architecture that can generalize well to the unseen target domain. 
In the following descriptions, let $\boldsymbol{\alpha}$, $\boldsymbol{\omega}$ and $\boldsymbol{\theta}_{G}$ denote the parameters for architecture topology, the supernet and the conditional generator $G(\cdot,\cdot)$, respectively. The conditional generator $G(\cdot,\cdot)$ takes data $\boldsymbol{x}$ and domain labels $\boldsymbol{\tilde{k}}$ as the input. Let $\ell_{\text{train}}$ be the training loss function, and $\ell_{\text{val}}$ be the validation loss function.

\noindent\textbf{Minimax optimization for NAS-OoD.} To generate a proper validation set for OoD generalization in NAS, as shown in Figure~\ref{fig:framework}, a conditional generator is learned to generate novel domain data by maximizing the losses on different neural architectures, while the optimal architecture variables are optimized by minimizing the losses on generated OoD images. This can be formulated as a constrained minimax optimization problem as follows:

\begin{equation}\label{eq:minmax}
\centering
\begin{aligned}
    &\mathop{\min}_{\boldsymbol{\alpha}} \mathop{\max}_{G}  \ell_{\text{val}}(\boldsymbol{\omega}^{\ast}(\boldsymbol{\alpha}), \boldsymbol{\alpha}, G(\boldsymbol{x},\boldsymbol{\tilde{k}})),\\ 
    &\text{s.t.} \quad
    \boldsymbol{\omega}^{\ast}(\boldsymbol{\alpha}) = \mathop{\arg \min}_{\boldsymbol{\omega}} \ell_{\text{train}}(\boldsymbol{\omega}, \boldsymbol{\alpha}, \boldsymbol{x}),
\end{aligned}
\end{equation}

where $G(\boldsymbol{x},\boldsymbol{\tilde{k}})$ is the generated data from the original input data $\boldsymbol{x}$ on the novel domain $\boldsymbol{\tilde{k}}$. This is different from NAS methods' formulation as we introduce a generator to adversarially generate challenging data from original input for validation loss to search for network architectures. This can avoid over-fitting problem by using the same data for optimizing neural network parameters and architectures as shown in our experiment. Solving this problem directly will involve calculating second-order derivatives that will bring much computational overhead and the constraint is hard to realize. Thus, we introduce the following practical implementations of our algorithm.

\begin{algorithm}[t]\small
\caption{NAS-OoD: Neural Architecture Search for OoD generalization}
\label{alg:method}
\begin{algorithmic}[1]
\REQUIRE Training set $\mathcal{D}$, batch size $n$, learning rate $\mu$. 
\ENSURE $\boldsymbol{\alpha}, \boldsymbol{\omega}, \boldsymbol{\theta}_G$.
\STATE Initialize $\boldsymbol{\alpha}$, $\boldsymbol{\omega}$, $\boldsymbol{\theta}_G$; 
\REPEAT
\STATE Sample a mini-batch of training images $\{(x_i, y_i)\}_{i=1}^{n}$;
\STATE Generate novel domain data: $x_i^{\text{syn}} \leftarrow \text{G} (x_i, \boldsymbol{\tilde{k}})$;
\STATE $\boldsymbol{\theta}_G \leftarrow \boldsymbol{\theta}_G - \mu \cdot \nabla_{\boldsymbol{\theta}_G} \ell_{\text{aux}}$ according to Eqn.~\eqref{eq:G};
\STATE $\boldsymbol{\omega} \leftarrow \boldsymbol{\omega} - \mu \cdot \nabla_{\boldsymbol{\omega}} \ell_{\text{train}}(\boldsymbol{\omega}, \boldsymbol{\alpha}, x_i)$ according to Eqn.~\eqref{eq:iterative};
\STATE $\boldsymbol{\theta}_G \leftarrow \boldsymbol{\theta}_G + \mu \cdot \nabla_{\boldsymbol{\theta}_G} \ell_{\text{val}}(\boldsymbol{\omega}, \boldsymbol{\alpha},x_i^{\text{syn}})$ according to Eqn.~\eqref{eq:iterative};
\STATE $\boldsymbol{\alpha} \leftarrow \boldsymbol{\alpha} - \mu \cdot \nabla_{\boldsymbol{\alpha}} \ell_{\text{val}}(\boldsymbol{\omega}, \boldsymbol{\alpha},x_i^{\text{syn}})$ according to Eqn.~\eqref{eq:iterative};
\UNTIL convergence;
\end{algorithmic}
\end{algorithm}

\noindent\textbf{Practical implementation.}
Inspired by the previous work in meta-learning~\cite{finn2017model}, we approximate the multi-step optimization with the one-step gradient when calculating the gradient for $\boldsymbol{\alpha}$. Different source domains are mixed together in architecture search, while the domain labels are embedded in the generator auxiliary loss training process which will be explained later. For the architecture search training process, architecture parameters $\boldsymbol{\alpha}$, network parameters $\boldsymbol{\omega}$ and parameters for conditional generator $\boldsymbol{\theta}_G$ are updated in an iterative training process: 
\begin{equation}\label{eq:iterative}
\begin{aligned}
&\boldsymbol{\omega} \leftarrow \boldsymbol{\omega} - \mu \cdot \nabla_{\boldsymbol{\omega}} \ell_{\text{train}}(\boldsymbol{\omega}, \boldsymbol{\alpha}, \boldsymbol{x}),\\
&\boldsymbol{\theta}_G \leftarrow \boldsymbol{\theta}_G + \mu \cdot \nabla_{\boldsymbol{\theta}_G} \ell_{\text{val}}(\boldsymbol{\omega}, \boldsymbol{\alpha}, G(\boldsymbol{x}, \boldsymbol{\tilde{k}})),\\
&\boldsymbol{\alpha} \leftarrow \boldsymbol{\alpha} - \mu \cdot \nabla_{\boldsymbol{\alpha}} \ell_{\text{val}}(\boldsymbol{\omega}, \boldsymbol{\alpha}, G(\boldsymbol{x},\boldsymbol{\tilde{k}})).
\end{aligned}
\end{equation}

\noindent\textbf{Generator’s auxiliary losses.} To train the generator and improve consistency, we apply an additional cycle consistency constraint to the generator:
\begin{equation}\label{eq:cycle}
\ell_{\text{cycle}} = ||\text{G}(\text{G}(\boldsymbol{x}_{k}, \widetilde{k}), k)-\boldsymbol{x}_{k}||_{1},
\end{equation}
where $\boldsymbol{x}_{k}$ denotes data from $K$ source domains with domain $\{s_1, s_2, ..., s_K\}$, $\widetilde{k}$ denotes the domain index for the generated novel domain $s_{K+1}$, and $||\cdot||_1$ refers to L1 norm. This can regularize the generator to be able to produce data from and back to the source domains.

To preserve semantic information, we also require the generated data $\text{G}(\boldsymbol{x}_{k}, \widetilde{k})$ to keep the same category as the original data $\boldsymbol{x}_{k}$.
\begin{equation}\label{eq:ce}
\ell_{\text{ce}}  = \text{CE}(Y(\text{G}(\boldsymbol{x}_{k}, \widetilde{k})),Y^{\ast}(\boldsymbol{x}_{k})),
\end{equation}
where $\text{CE}$ be the cross-entropy loss, $Y$ is a classifier with a few convolutional layers pretrained on training data, $Y^{\ast}(\cdot)$ is the ground-truth labels for the input data.

The total auxiliary loss for generator is defined as follows:
\begin{equation}\label{eq:G}
\ell_{\text{aux}} = \lambda_{cycle} \cdot \ell_{\text{cycle}} + \lambda_{ce} \cdot \ell_{\text{ce}},
\ell_{\text{ot}},
\end{equation}
where $\lambda_{ce}$ and $\lambda_{cycle}$ 
are hyper-parameters.

Compared with the gradient-based perturbation~\cite{shankar2018generalizing}, the conditional generator is able to model a more sophisticated distribution shift due to its intrinsic learnable nature.
The NAS-OoD algorithm is outlined in Algorithm ~\ref{alg:method}.

\section{Experiments}\label{sec:expe}

In this section, we conduct numerical experiments to evaluate the effectiveness of our proposed NAS-OoD method. 
To provide a comprehensive comparison with baselines, We compare our proposed NAS-OoD with the SOTA algorithms from various OoD areas, including empirical risk minimization (ERM~\cite{arjovsky2019invariant}),
invariant risk minimization (IRM~\cite{arjovsky2019invariant}),
risk extrapolation (REx~\cite{krueger2020outofdistribution}),
domain generalization by solving jigsaw puzzles (JiGen~\cite{carlucci2019domain}),
mixup (Mixup~\cite{zhang2017mixup}),
curriculum mixup (Cumix~\cite{mancini2020towards}),
marginal transfer learning (MTL~\cite{blanchard2017domain}),
domain adversarial training (DANN~\cite{ganin2016domain}), 
correlation alignment (CORAL~\cite{sun2016deep}),
maximum mean discrepancy (MMD~\cite{li2018domain}),
distributionally robust neural network (DRO~\cite{sagawa2019distributionally}),
convnets with batch balancing (CNBB~\cite{he2020towards}),
cross-gradient training (CrossGrad~\cite{shankar2018generalizing}), and the recently proposed
decomposed feature representation and semantic augmentation (DecAug~\cite{bai2020decaug}), etc. 
More details about the baseline methods are shown in the Appendix.

For ablation studies, We also compare NAS-OoD with SOTA NAS methods, such as differentiable architecture search (DARTS~\cite{liu2018darts}),
stochastic neural architecture search (SNAS~\cite{xie2018snas}),
efficient and consistent neural architecture search by sparse coding (ISTA-NAS~\cite{yang2020ista}). This is to test whether naively combining NAS methods with OoD learning algorithms can improve the generalization performance.

Our framework was implemented with PyTorch 1.4.0 and CUDA v9.0. We conducted experiments on NVIDIA Tesla V100. 
Following the design of~\cite{choi2018stargan}, our generator model has an encoder-decoder structure, which consists of two down-sampling convolution layers with stride 2, three residual blocks, and two transposed convolution layers with stride 2 for up-sampling. The domain indicator is encoded as a one-hot vector. The one-hot vector is first spatially expanded and then concatenated with the input image to train the generator. More implementation details can be found in the Appendix.

\subsection{Evaluation Datasets}
\label{exp:dataset}

We evaluate our NAS-OoD on four challenging OoD datasets: NICO Animal, NICO Vehicle, PACS, and Out-of-Distribution Fingerprint (OoD-FP) dataset where methods have to be able to perform well on data distributions different from training data distributions. The evaluation metric is the classification accuracy of the test set. The number of neural network parameters is used to measure the computational complexity for comparison between different neural network architectures.

\begin{figure}[t]
    \centering
    \subfigure[In water]{\includegraphics[height = 3.2cm]{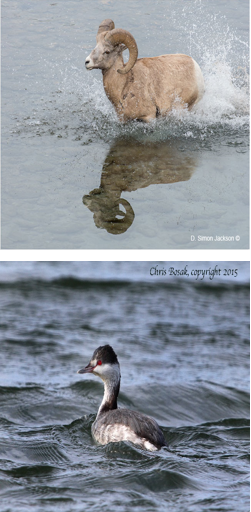}}  
    \subfigure[On snow]{\includegraphics[height = 3.2cm]{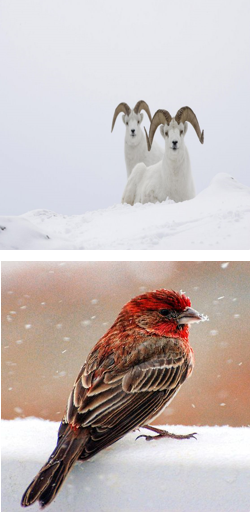}}
    \subfigure[On grass]{\includegraphics[height = 3.2cm]{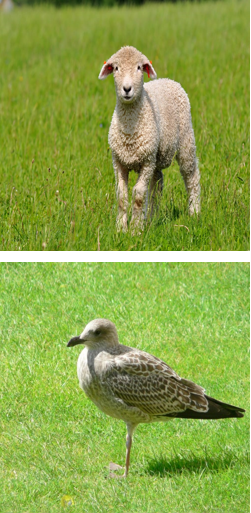}}
    \hspace{1mm}
    \subfigure[Others]{\includegraphics[height = 3.2cm]{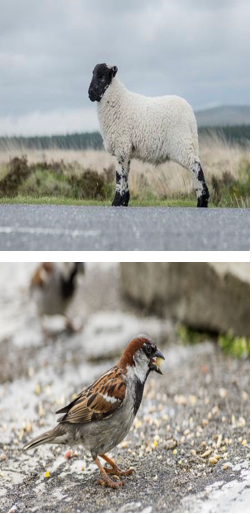}}
    \caption{Examples of the out-of-distribution data from the NICO dataset with contexts (a) In water, (b) On snow, (c) On grass, and(d) Others.}
    \vspace{-0.2cm}
    \label{fig:imgnico}
\end{figure}

\begin{figure}[t]
    \centering
    \subfigure[P]{\includegraphics[height = 3.15cm]{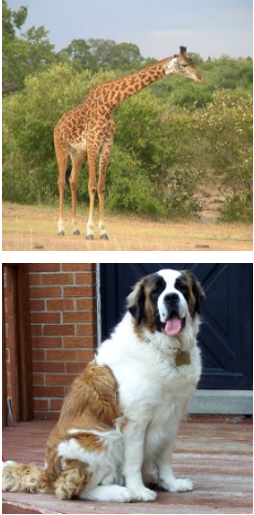}}
    \subfigure[A]{\includegraphics[height = 3.15cm]{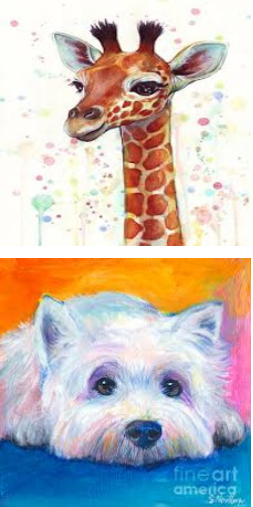}}
    \subfigure[C]{\includegraphics[height = 3.15cm]{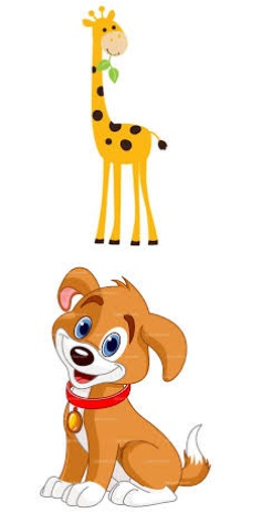}}
    \subfigure[S]{\includegraphics[height = 3.15cm]{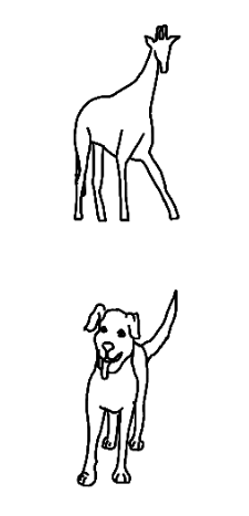}}
    \caption{Typical examples of out-of-distribution data with diversity shift from the PACS dataset. (a) Photo. (b) Art Painting. (c) Cartoon. (d) Sketch. 
    }
\label{fig:imgpacs}
\end{figure}

\begin{figure}[t]
    \centering
    \subfigure[Domain 1]{\includegraphics[height = 3.15cm]{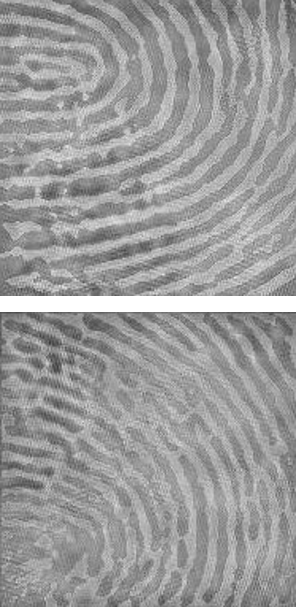}}
    \hspace{3mm}
    \subfigure[Domain 2]{\includegraphics[height = 3.15cm]{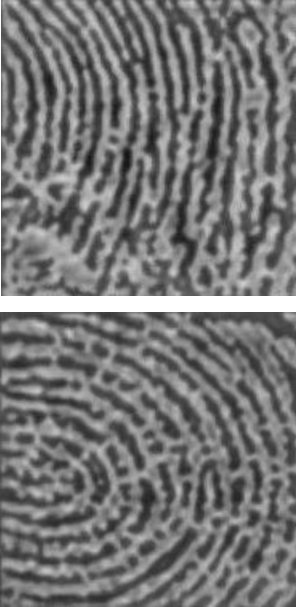}}
    \hspace{3mm}
    \subfigure[Domain 3]{\includegraphics[height = 3.15cm]{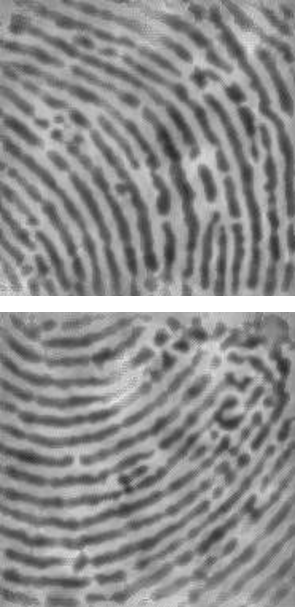}}
    \caption{Typical examples of out-of-distribution data in the OoD-FP dataset with three different domains.}
    \label{fig:imgFP}
\end{figure}

\noindent\textbf{NICO (Non-I.I.D. Image with Contexts) dataset:} NICO consists of two datasets, i.e., NICO Animal with 10 classes and NICO Vehicle with 9 classes. 
The NICO dataset is a recently proposed OoD generalization dataset in the real scenarios~\cite{he2020towards} (see Figure~\ref{fig:imgnico}), which contains different contexts, such as different object poses, positions, and backgrounds across the training, validation, and test sets.

\noindent\textbf{PACS (Photo, Art painting, Cartoon, Sketch) dataset:} This dataset is commonly used in OoD generalization (see Figure~\ref{fig:imgpacs}). It contains four domains with different image styles, namely photo, art painting, cartoon, and sketch with seven categories (dog, elephant, giraffe, guitar, horse, house, person). We follow the same leave-one-domain-out validation experimental protocol as in~\cite{Li2017}, i.e., we select three domains for training and the remaining domain for testing for each time.

\noindent\textbf{OoD-FP (Out-of-Distribution Fingerprint) dataset:} The OoD-FP dataset is a real industry dataset that contains three domains corresponding to different fingerprint collection devices on different brands of mobile phones (see Figure~\ref{fig:imgFP}). In the fingerprint recognition task, the goal is to learn to distinguish whether input fingerprints are from the users' fingerprints stored in the dataset. Due to the hardware implementation differences, fingerprints exhibit different styles from different devices. In our setting, the goal is to learn a universal fingerprint recognition neural network to generalize on the fingerprints collected from unseen datasets.

\begin{table}[t]
\centering
\caption{Results of NAS-OoD compared with different methods with ResNet-18 (11.7M) on the NICO dataset.}
\label{table:nico}
\begin{adjustbox}{max width=0.95\textwidth}
\begin{threeparttable}
\begin{tabular}{lcc|c}
\toprule
\toprule
Model & Animal & Vehicle & Average \\
\midrule
ERM~\cite{arjovsky2019invariant}	               &75.87   &74.52   &75.19  \\
IRM~\cite{arjovsky2019invariant}	               &59.17   &62.00   &60.58	 \\
REx~\cite{krueger2020outofdistribution}            &74.31   &66.20   &70.25  \\
JiGen~\cite{carlucci2019domain}                    &84.95   &79.45   &82.20  \\
Mixup~\cite{zhang2017mixup}\tnote{*}	           &80.27   &77.00   &78.63	 \\
Cumix~\cite{mancini2020towards}                    &76.78   &74.74   &75.76  \\
MTL~\cite{blanchard2017domain}\tnote{*}            &78.89   &75.11   &77.00	 \\
DANN~\cite{ganin2016domain}                        &75.59   &72.23   &73.91  \\
CORAL~\cite{sun2016deep}\tnote{*}                  &80.27   &71.64   &75.95	 \\
MMD~\cite{li2018domain}\tnote{*}                   &70.91   &68.04   &69.47  \\
DRO~\cite{sagawa2019distributionally}\tnote{*}     &77.61   &74.59   &76.10  \\
CNBB~\cite{he2020towards}                          &78.16   &77.39   &77.77  \\
DecAug~\cite{bai2020decaug}                        &85.23   &80.12   &82.67  \\
\midrule
\emph{NAS-OoD} &\emph{\textbf{88.72}} &\emph{\textbf{81.59}} &\emph{\textbf{85.16}}\\
\midrule
Params (M) & 3.25 & 3.00 & 3.13
\\
\bottomrule
\bottomrule
\end{tabular}
\begin{tablenotes}
	\item[*] Implemented by ourselves.
\end{tablenotes}
\end{threeparttable}
\end{adjustbox}
\vspace{-10pt}
\end{table}

\subsection{Results and Discussion}
\label{exp:discuss}
NAS-OoD achieves the SOTA performance \emph{simultaneously} on various datasets from different OoD research areas, such as stable learning, domain generalization, and real industry dataset.

\begin{table}[t]
\centering
\caption{Classification accuracy on the PACS dataset compared with different methods with ResNet-18 (11.7M).}
\label{table:pacs_pretrain}
\begin{adjustbox}{max width=0.48\textwidth}
\begin{threeparttable}
\begin{tabular}{lcccc|c}
\toprule
\toprule
Model  &A  &C  &S  &P  &Average\\
\midrule
ERM~\cite{arjovsky2019invariant}              &77.85   &74.86   &67.74   &95.73   &79.05  \\
IRM~\cite{arjovsky2019invariant}              &70.31   &73.12   &75.51   &84.73   &75.92  \\
REx~\cite{krueger2020outofdistribution}       &76.22   &73.76   &66.00   &95.21   &77.80  \\
JiGen~\cite{carlucci2019domain}               &79.42   &75.25   &71.35   &96.03   &80.51  \\
Mixup~\cite{zhang2017mixup}\tnote{*}          &82.01   &72.58   &72.48   &93.29   &80.09  \\
CuMix~\cite{mancini2020towards}               &82.30   &76.50   &72.60   &95.10   &81.60  \\
MTL~\cite{blanchard2017domain}\tnote{*}       &76.76   &71.87   &76.73   &92.65   &79.50  \\
MLDG~\cite{li2018learning}                    &79.50   &77.30   &71.50   &94.30   &80.70  \\
MASF~\cite{dou2019domain}                     &80.29   &77.17   &71.69   &94.99   &81.03  \\
DANN~\cite{ganin2016domain}                   &81.30   &73.80   &74.30   &94.00   &80.80  \\
CORAL~\cite{sun2016deep}\tnote{*}             &80.49   &74.32   &75.06   &94.09   &80.99  \\
MMD~\cite{li2018domain}\tnote{*}	          &79.34   &73.76   &72.61   &94.19   &79.97  \\
DRO~\cite{sagawa2019distributionally}\tnote{*}&78.09   &74.18   &77.00   &93.45   &80.68  \\
CrossGrad~\cite{shankar2018generalizing}      &78.70   &73.30   &65.10   &94.00   &80.70  \\
L2A-OT~\cite{zhou2020learning} &83.30 &78.20 &73.60 &96.20 &82.80 \\
DecAug~\cite{bai2020decaug}                   &79.00   &79.61   &75.64   &95.33   &82.39  \\
\midrule
\emph{NAS-OoD} &\emph{\textbf{83.74}} &\emph{\textbf{79.69}} &\emph{\textbf{77.27}} &\emph{\textbf{96.23}} &\emph{\textbf{84.23}} \\
\midrule
Params (M) & 3.51 & 3.44 & 3.35 & 3.15 & 3.36 \\
\bottomrule
\bottomrule
\end{tabular}
\begin{tablenotes}
	\item[*] Implemented by ourselves.
\end{tablenotes}
\end{threeparttable}
\end{adjustbox}
\vspace{-10pt}
\end{table}

The results for the challenging NICO dataset are shown in Table~\ref{table:nico}. From Table~\ref{table:nico}, the proposed NAS-OoD method achieves the SOTA performance simultaneously on the two subsets of the NICO dataset with a much fewer number of parameters. Specifically, NAS-OoD achieves $88.72\%$ on NICO Animal and $81.59\%$ on NICO Vehicle with only around $3.1$ million parameters compared with DecAug achieving $82.67\%$ accuracy but with $11.7$ million parameters. The superior performance of NAS-OoD also confirms the possibility of improving the neural network's OoD generalization performance by searching for the architecture, which provides an orthogonal way to improve the OoD generalization.

We also compare our methods with different domain generalization methods on the PACS dataset. The results are shown in Table~\ref{table:pacs_pretrain}. Similarly, we observe that NAS-OoD achieves SOTA performance on all the four domains and the best average generalization performance of $83.89\%$ with only $3.36$ million of network parameters. The generalization accuracy is much better than previous OoD algorithms DecAug ($82.39\%$), JiGen ($80.51\%$), IRM ($75.92\%$) with ResNet-18 backbone, which are the best OoD approaches before NAS-OoD. The network parameters for ResNet-18 is $11.7$ million which is much larger than the network searched by our NAS-OoD. Note that the relative performance for some algorithms may change drastically between NICO and PACS datasets whereas the proposed NAS-OoD algorithm can generalize well simultaneously on datasets from different OoD research areas.

To test the generalization performance of NAS-OoD on real industry datasets, we compare NAS-OoD with other methods on OoD-FP dataset. The results are shown in Table~\ref{table:finger}. NAS-OoD consistently achieves good generalization performance with the non-trivial improvement compared with other methods. NAS-OoD achieves a $1.23\%$ error rate in the fingerprint classification task which almost reduces the error rate by around $70\%$ compared with the second-best method--MMD. This demonstrates the superiority of NAS-OoD and especially its potential to be practically useful in real industrial applications.

\begin{table}[t]
\centering
\caption{Classification accuracy compared to different methods with ResNet-18 backbone (11.7M) on the OoD-FP dataset. All methods are implemented by ourselves.
}
\label{table:finger}
\begin{adjustbox}{max width=0.48\textwidth}
\begin{tabular}{lccc|c}
\toprule
\toprule
Model  & Domain 1  & Domain 2  & Domain 3   & Average\\
\midrule
ERM~\cite{arjovsky2019invariant}        &93.75  &92.70  &92.70  &93.05  \\
IRM~\cite{arjovsky2019invariant}        &95.83  &87.50  &84.37  &89.23  \\
REx~\cite{krueger2020outofdistribution} &97.91  &91.66	&92.70  &94.09  \\
Mixup~\cite{zhang2017mixup}             &96.87  &97.91  &90.62	&95.13  \\
MTL~\cite{blanchard2017domain}          &95.83	&97.91  &90.62  &94.78  \\
DANN~\cite{ganin2016domain}             &95.83	&97.91	&86.45  &93.39  \\
CORAL~\cite{sun2016deep}                &94.79  &97.91  &91.66	&94.78  \\
MMD~\cite{li2018domain}              	&96.87  &95.83  &94.79	&95.83  \\
DRO~\cite{sagawa2019distributionally}   &96.87  &95.83  &89.58	&94.09  \\
\midrule
\emph{NAS-OoD} &\emph{\textbf{99.27}}  &\emph{\textbf{99.49}} &\emph{\textbf{97.54}} &\emph{\textbf{98.77}} 
\\
\midrule
Params (M) & 2.28 & 2.28 & 2.43 & 2.33
\\
\bottomrule
\bottomrule
\end{tabular}
\end{adjustbox}
\vspace{-10pt}
\end{table}

\subsection{Ablation Study}
\label{exp:abla}

In this section, we first test whether naively combining NAS methods with domain generalization methods can achieve good OoD generalization performances. We conduct experiments on the NICO dataset. 
The results are shown in Table~\ref{table:niconas}. It can be seen that using NAS methods only, such as DARTS, can achieve only $79.61\%$ average accuracy, significantly lower than most compared compositions. This is in stark contrast with the good performance for NAS methods on IID generalization tasks where training and test datasets have similar distributions. This is because NAS methods are doing variational optimization by finding not only the best parameters but also the best function for fitting whereas this can help NAS methods to achieve good performance in IID settings. In OoD settings, where test data distributions differ significantly from training data distributions, NAS methods can overfit the training data distribution and achieve sub-optimal performance. Besides, we can also observe that naively combining the NAS methods with OoD learning algorithms, such as IRM, brings no statistically significant performance gain. The reason is that many OoD learning algorithms are based on implicit or explicit regularization added to the ERM loss. NAS methods will explore the search space to fit the loss terms and the regularization term may be ignored as NAS methods may exploit only the ERM loss, thus bringing no performance gain. This also confirms that generating OoD data is needed during training to avoid over-fitting.

As shown in Table~\ref{table:variants}, the average results of randomly sampled architectures are 79.45\% (Animal) and 75.70\% (Vehicle), which is significantly lower than those of the architectures searched by NAS-OoD. We also conduct an ablation study on this auxiliary loss, the average accuracy on NICO without cycle loss is 83.86\%, which is low than our proposed method 85.16\%. This shows the effectiveness of the auxiliary cycle loss to facilitate the searching process.

\subsection{Analysis of Searched Architectures}
After setting up the NAS-OoD framework, we want to analyze whether any special patterns for searched cell-based network architectures and whether the NAS-OoD framework can stably find consistent architectures.

\begin{table}[t]
\centering
\caption{Results of NAS-OoD compared with other NAS methods. The baselines are implemented by ourselves.}
\label{table:niconas}
\begin{adjustbox}{max width=0.95\textwidth}
\begin{threeparttable}
\begin{tabular}{lcc|c}
\toprule
\toprule
Model &Animal &Vehicle &Average \\
\midrule
DARTS~\cite{liu2018darts}  & 83.67  &75.55 & 79.61 \\
DARTS + IRM~\cite{liu2018darts}	& 82.29 & 72.24 & 77.26 \\
SNAS~\cite{xie2018snas}  & 85.96& 80.04 & 83.00 \\
SNAS + IRM~\cite{xie2018snas}	 & 82.94 & 76.51 & 79.73 \\
ISTA-NAS~\cite{yang2020ista} & 86.70 & 80.56 & 83.63\\
ISTA-NAS + IRM~\cite{yang2020ista} & 82.57 & 77.61 & 80.09 \\
\midrule
\emph{NAS-OoD} & \emph{\textbf{88.72}} & \emph{\textbf{81.59}}&\emph{\textbf{85.16}}\\
\bottomrule
\bottomrule
\end{tabular}
\end{threeparttable}
\end{adjustbox}
\end{table}

\begin{table}[t]
\centering
\caption{NAS-OoD variants.}
\label{table:variants}
\begin{adjustbox}{max width=0.9\textwidth}
\begin{threeparttable}
\begin{tabular}{lcc|c}
\toprule
\toprule
Model &Animal &Vehicle &Average \\
\midrule
Random Sample &80.92 &76.43 &78.68 \\
NAS-OoD w/o cycle loss     &86.88 &80.85 &83.86 \\
\midrule
\emph{NAS-OoD} & \emph{\textbf{88.72}} & \emph{\textbf{81.59}}&\emph{\textbf{85.16}}\\
\bottomrule
\bottomrule
\end{tabular}
\end{threeparttable}
\end{adjustbox}
\end{table}

\begin{figure}
    \centering
    \includegraphics[width=0.7\linewidth]{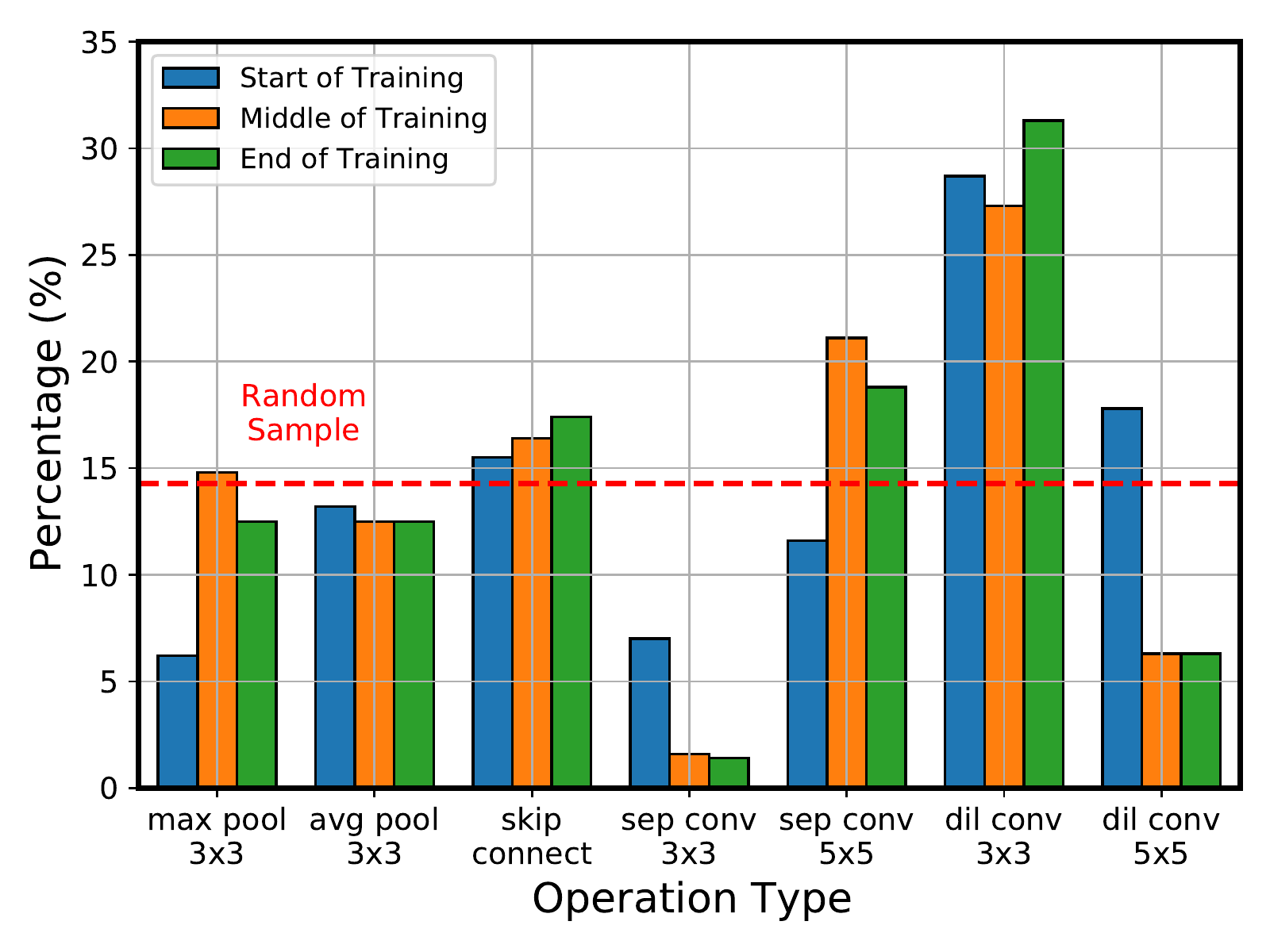}
    \caption{Temporal stability of search architecture.(Better viewed in the zoom-in mode)}
    \label{fig:multiprocess}
\end{figure}

\noindent \textbf{Temporal stability of searched architecture}
To check whether the found special pattern is consistent during the training process, we plot the operation type's percentage during the training process in Figure~\ref{fig:multiprocess}. In Figure~\ref{fig:multiprocess}, we can find that as the training proceeds, the architecture found by NAS-OoD is converging to the pattern that the percentage of dilated convolution $3\times3$ is higher and the separable convolution $3\times3$ is lower. This might be because dilated convolution has a larger receptive field than separable convolution which only receives one channel for each convolution kernel and a large receptive field can better learn the shape features of objects rather than the spurious features, such as color and texture.

\begin{figure}
    \centering
    \includegraphics[width=0.7\linewidth]{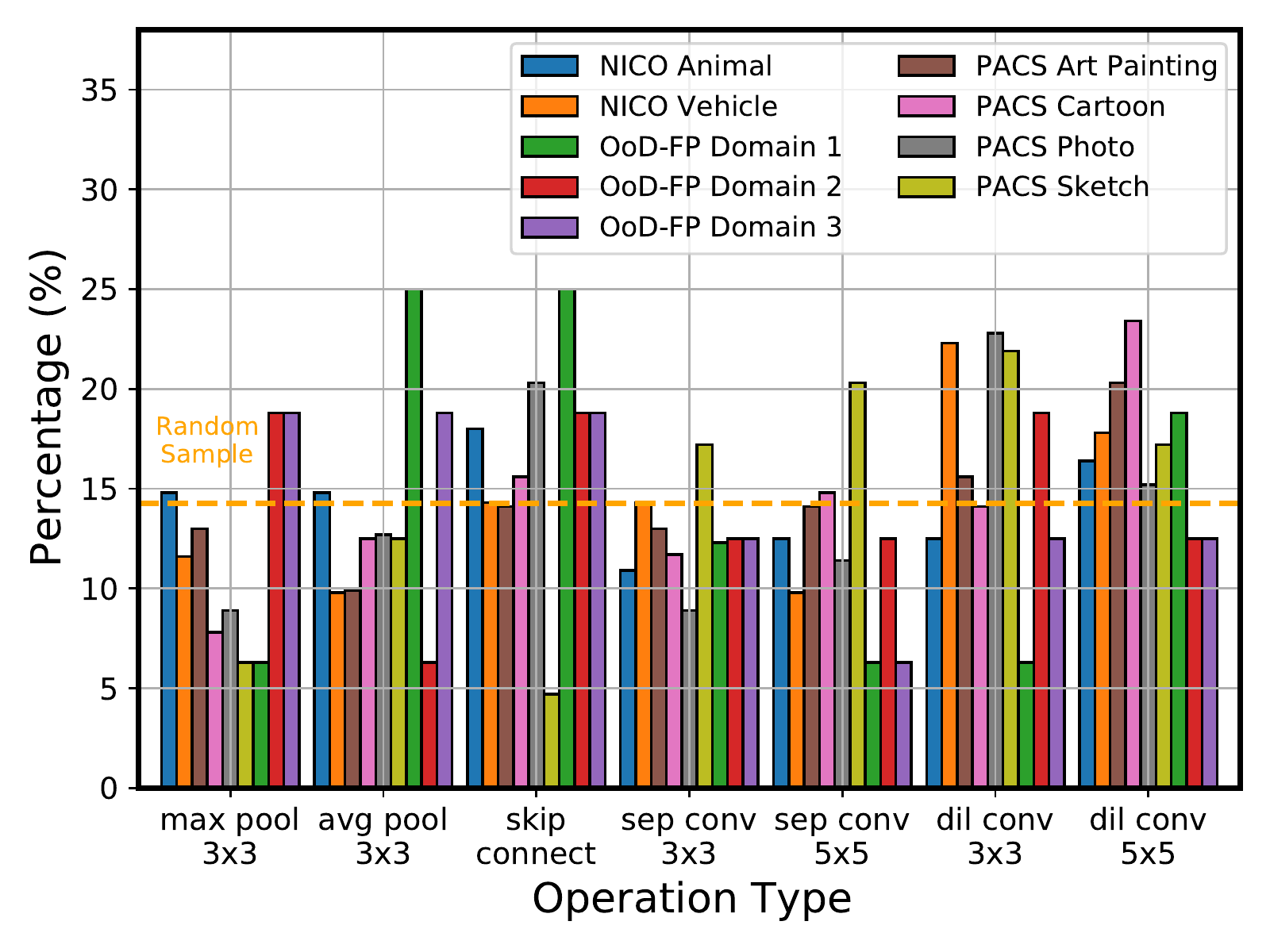}
    \caption{Statistical analysis of searched architectures on different datasets.(Better viewed in the zoom-in mode)}
    \label{fig:multidatasets}
\end{figure}

\noindent\textbf{Cross dataset architecture patterns}
To check whether the architecture patterns searched by NAS-OoD are similar across different datasets, we plot the operation type's percentage for different datasets in Figure~\ref{fig:multidatasets}.

We found there are similarities of architectures found on different datasets. Specifically, the searched NAS-OoD architectures tend to contain more convolutional operations with a large kernel size compared with randomly sampled architectures. This may be because a larger kernel size convolution operation has larger receptive fields, which makes better use of contextual information compared with a small kernel size. NAS-OoD architectures also present more skip connection operations compared with random selection and locate on both skip edges and direct edges, which can better leverage both low-level texture features and high-level semantic information for recognition. 
There is some previous study show that densely connected pattern benefits model robustness.
NAS-OoD searched for more dilated convolutions than normal convolutions, which may be due to that the dilated convolutions enlarge the receptive fields.

\begin{figure}
    \centering
    \subfigure[Normal cells]{
    \includegraphics[height = 0.285\linewidth]{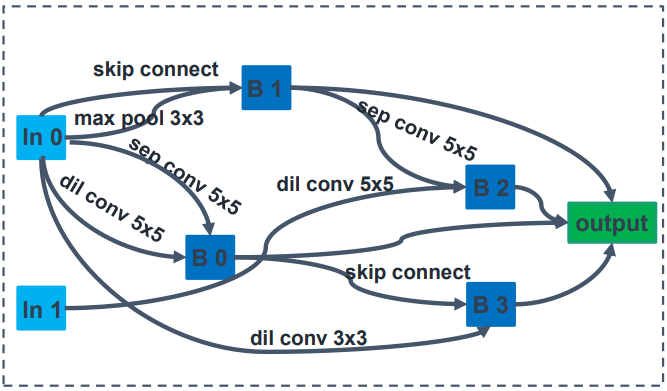}}
    \subfigure[Reduce cells]{\includegraphics[height = 0.285\linewidth]{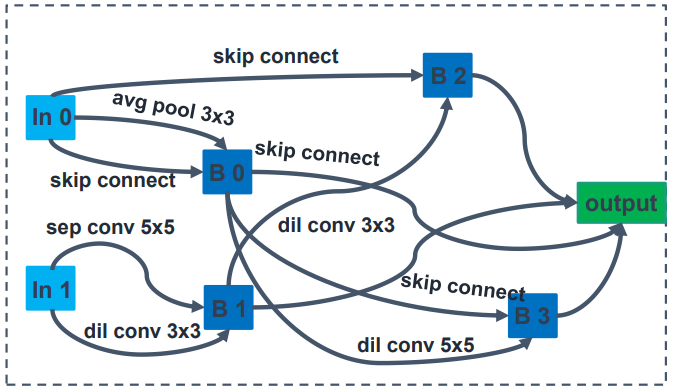}}
    \caption{Typical examples of searched robust architectures on NICO dataset.(Better viewed in the zoom-in mode)}
    \label{fig:nicocell}
\end{figure}

\noindent\textbf{Visualization of the searched architecture cells.}
As illustrated in Figure~\ref{fig:nicocell}, we present the detailed structures of the best cells discovered on different datasets using NAS-OoD. Figure~\ref{fig:nicocell} (a) show the normal cells and (b) demonstrate the reduce cells. The searched cell contains two input nodes, four intermediate nodes, and one output node. Each intermediate node has two edges to the previous nodes which consist of both direct edge and skip edge, and the operation is presented on each edge. Besides, all the intermediate nodes are connected and aggregated to the output node.

\begin{figure}
    \centering
    \includegraphics[width=0.98\linewidth]{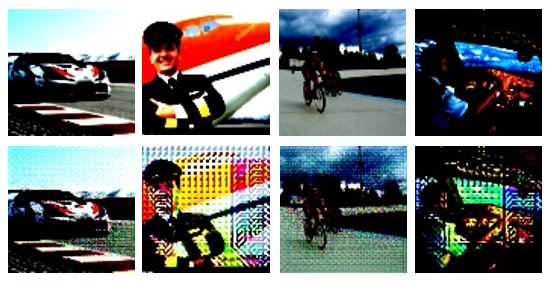}
    \caption{Some examples of synthetic images. The first row shows the original images, and the second row is its corresponding synthetic images.}
    \label{fig:syn}
\end{figure}

\noindent\textbf{Visualization of generated OoD data.}
We also visualize the generated OoD data in Figure~\ref{fig:syn}. We find that the generated images show different properties and are clearly different from the source images. The conditional generator tends to generate images with different background patterns, textures, and colors. The semantic different make them helpful for improving out-of-distribution generalization ability.

\section{Conclusions} \label{sec:conc}

We propose a robust neural architecture search framework that is based on differentiable NAS to understand the importance of network architecture against Out-of-Distribution robustness. 
We jointly optimize NAS and a conditional generator in an end-to-end manner. The generator is learned to synthesize OoD instances by maximizing their losses computed by different neural architectures, while the goal for architecture search is to find the optimal architecture parameters that minimize the synthesized OoD data losses. 
Our study presents several valuable observations on designing robust network architectures for OoD generalization. Extensive experiments show the effectiveness of NAS-OoD, achieving state-of-the-art performance on different OoD datasets with discovered architectures having a much fewer number of parameters.


\section*{Acknowledgements}
\noindent
This work was supported, in part, by Hong Kong General Research Fund (under grant number 16200120)

\clearpage

{\small
\bibliographystyle{ieee_fullname}
\bibliography{main}
}

\clearpage

\appendix

\begin{spacing}{1.1}



\section{Cross Check of Discovered Architectures}
To check whether the architecture discovered by NAS-OoD can perform well across datasets, we present the OoD performance on different datasets in Table~\ref{table:cross}. The architectures searched on PACS and NICO are finetuned and evaluated on OoD-FP.

\begin{table}[h]
\centering
\caption{Classification accuracy of the discovered architectures using NAS-OoD cross-check on different datasets.
}
\label{table:cross}
\begin{adjustbox}{max width=0.45\textwidth}
\begin{tabular}{lccc|c}
\toprule
\toprule
Model         &Domain 1 &Domain 2 &Domain 3 &Average\\
\midrule
PACS (Art Painting) & 99.10 & 98.87 & 96.55 & 98.17 \\
PACS (Cartoon)      & 99.35 & 98.86 & 94.82 & 97.68 \\
PACS (Sketch)       & 99.07 & 98.23 & 96.98 & 98.09 \\
PACS (Photo)        & \emph{\textbf{99.45}} & 99.06 & 96.68 & 98.40 \\
NICO Animal       & 98.53 & 97.55 & 87.16 & 94.41 \\
NICO Vehicle      & 96.33 & 90.16 & 85.70 & 90.73 \\
\midrule
\emph{NAS-OoD}    & 99.27 & \emph{\textbf{99.49}} & \emph{\textbf{97.54}} & \emph{\textbf{98.77}}
\\
\bottomrule
\bottomrule
\end{tabular}
\end{adjustbox}
\end{table}

As shown in Table~\ref{table:cross}, we find that the architectures discovered by NAS-OoD on PACS and NICO also show good performance on OoD-FP. Architectures discovered on PACS (Photo) can achieve 98.40\% average accuracy, significantly higher than other OoD algorithms on OoD-FP.
This may be because the architectures found by NAS-OoD contain some common patterns such as larger kernel size and more percentage of dilated convolutions that can better make use of contextual information and shows better OoD generalization ability. It can be seen that the architecture discovered on NICO Animal achieves 94.41\% on OoD-FP with 3.25 million parameters, while the architecture discovered on NICO Vehicle achieves 90.73\% on OoD-FP with 3.00 million parameters.
This might be because that the architecture discovered with a higher number of parameters may achieve better OoD generalization performance.

\section{More Ablation Study Results}

We present more ablation results to test whether directly combining NAS with domain generalization methods can achieve good OoD generalization performance or not. The results on the PACS dataset are shown in Table~\ref{table:pacs-nasdg}.

\begin{table}[h]
\centering
\caption{Results of NAS-OoD variants on PACS dataset.}
\label{table:pacs-nasdg}
\begin{adjustbox}{max width=0.45\textwidth}
\begin{threeparttable}
\begin{tabular}{lcccc|c}
\toprule
\toprule
Model  &A  &C  &S  &P  &Average\\
\midrule
NAS + IRM  &79.59  &75.34  &65.11 &95.69 &78.93  \\
NAS + REx  &80.86 &75.68 &73.94 &96.05 &81.63 \\
\midrule
\emph{NAS-OoD} &\emph{\textbf{82.42}} &\emph{\textbf{79.69}} &\emph{\textbf{77.27}} &\emph{\textbf{96.17}} &\emph{\textbf{83.89}} \\
\bottomrule
\bottomrule
\end{tabular}
\end{threeparttable}
\end{adjustbox}
\vspace{-10pt}
\end{table}

In Table~\ref{table:pacs-nasdg}, we test the performance of NAS methods with different OoD generalization algorithms on the PACS dataset (Table~\ref{table:pacs-nasdg}) and NICO dataset (Table~\ref{table:nico-nasdg}). We can observe that directly combining the NAS methods with different OoD algorithms, such as IRM (78.93\%) and REx (81.63\%), achieves lower performance than NAS-OoD (83.89\%).
This is expected since the regularization terms combined with NAS methods are computed on the original training data that cannot provide the correct supervisory signal for architecture search.
The results confirm that jointly optimizing the data generator and the neural network architecture is needed to discover robust architectures against distribution shifts.

\begin{table}[t]
\centering
\caption{More results of NAS-OoD variants on NICO dataset. The baselines are implemented by ourselves.}
\label{table:nico-nasdg}
\begin{adjustbox}{max width=0.45\textwidth}
\begin{threeparttable}
\begin{tabular}{lcc|c}
\toprule
\toprule
Model &Animal &Vehicle &Average \\
\midrule
DARTS + IRM      &82.29 &72.24 &77.26 \\
DARTS + REx      &79.17 &77.39 &78.28 \\
SNAS + IRM       &82.94 &76.51 &79.73 \\
SNAS + REx       &73.30 &69.15 &71.23 \\
ISTA-NAS + IRM   &82.57 &77.61 &80.09 \\
ISTA-NAS + REx   &73.71 &69.80 &71.76 \\
\midrule
\emph{NAS-OoD} & \emph{\textbf{88.72}} & \emph{\textbf{81.59}}&\emph{\textbf{85.16}}\\
\bottomrule
\bottomrule
\end{tabular}
\end{threeparttable}
\end{adjustbox}
\end{table}

\section{Experimental Results on PACS datasets}

\begin{table}[h]
\centering
\caption{Classification accuracy on the PACS dataset (train from scratch). The backbone for the baselines is ResNet-18 (11.7M).}
\label{table:pacs-scratch}
\begin{adjustbox}{max width=0.48\textwidth}
\begin{threeparttable}
\begin{tabular}{lcccc|c}
\toprule
\toprule
Model  &A  &C  &S  &P  &Average\\
\midrule
ERM~\cite{arjovsky2019invariant}\tnote{*}        &33.11 &42.92 &40.39 &53.83 &42.56 \\
IRM~\cite{arjovsky2019invariant}\tnote{*}        &30.08 &41.85 &35.56 &39.10 &36.65 \\
REx~\cite{krueger2020outofdistribution}\tnote{*} &31.93 &45.95 &35.84 &44.19 &39.48 \\
Mixup~\cite{zhang2017mixup}\tnote{*}             &35.16 &47.87 &42.12 &53.59 &44.69 \\
MTL~\cite{blanchard2017domain}\tnote{*}          &38.48 & 49.06 &46.55 &51.98 &46.52 \\
DANN~\cite{ganin2016domain}\tnote{*}             &32.71 &48.76 &40.88 &54.79 &44.29 \\
CORAL~\cite{sun2016deep}\tnote{*}                &42.63 &49.96 &46.78 &56.29 &48.92  \\
MMD~\cite{li2018domain}\tnote{*}	             &39.89 &51.11 &43.09 &53.41 &46.88 \\
DRO~\cite{sagawa2019distributionally}\tnote{*}   &38.92 &46.72 &46.73  &54.67  &46.76  \\
DecAug~\cite{bai2020decaug}\tnote{*}             &45.51 &54.22 &45.05 &59.04 &50.96 \\
\midrule
\emph{NAS-OoD} & \emph{\textbf{48.24}} & \emph{\textbf{54.27}} & \emph{\textbf{46.93}} & \emph{\textbf{65.75}} & \emph{\textbf{53.80}} \\
\midrule
Param (M) & 3.51 & 3.44 & 3.35 & 3.15 & 3.36 \\
\bottomrule
\bottomrule
\end{tabular}
\begin{tablenotes}
    \item[*] Implemented by ourselves.
\end{tablenotes}
\end{threeparttable}
\end{adjustbox}
\end{table}

We present more experimental results on PACS datasets. The methods are trained from scratch, and the backbone for the baselines is ResNet-18. As shown in Table~\ref{table:pacs-scratch}, we observe that the architecture searched by NAS-OoD achieves SOTA performance on PACS dataset. The generalization accuracy is much better than previous OoD algorithms. This further demonstrates the superiority of NAS-OoD.

\section{Architectures Discovered on ImageNet}

More recent architectures discovered on ImageNet like EfficientNet-B0 (5.3M) achieves 82.66\% on NICO Animal and 78.57\% on NICO Vehicle, which is lower than architectures searched by NAS-OoD. EfficientNet is searched by traditional NAS methods focusing on IID settings which can easily overfit the training data distribution.

\section{More Recent Data Augmentation Methods}

We tried other data augmentation methods RandAugment (NeurIPS 2020) and obtained 77.84\% on NICO Vehicle, which is lower than NAS-OoD. Instead of just adding noise to images, our proposed method generates worst-case OoD samples with semantic meanings automatically, which helps train a robust classifier.

\section{NAS-OoD in the Standard Setting}

Compared with Top1 accuracy 93.02\% obtained by ResNet-18 (11.7M) from He et al. (CVPR 2016), our method achieves Top1 accuracy 95.28\% on standard datasets CIFAR-10 without extra training data and under low model capacity (3.32M). This shows that our method also works well in standard settings without domain shift.

\section{T-SNE of Architecture Topology}

\begin{figure}[h]
    \centering
    \includegraphics[width=0.6\linewidth]{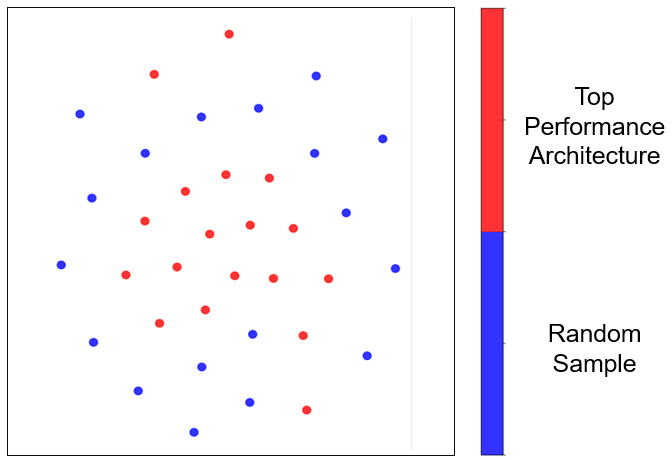}
    \caption{t-SNE visualization of architecture topology.}
    \label{fig:t-SNE}
\end{figure}

We do t-SNE visualization to visualize the embedding of the topology of $\alpha$ in the two-dimensional space to see whether the top performance architecture has different patterns compared with average architectures. We randomly selected nine top performance architectures found by NAS-OoD and compared them with other randomly selected architectures. The results are shown in Figure~\ref{fig:t-SNE}. The top performance architecture is significantly different from random architectures. This demonstrates that there are indeed special architecture patterns that can help the deep neural network to generalize well on OoD settings.

\section{Implementation Details}

We conduct a fair comparison of NAS-OoD with various OoD generalization methods and SOTA NAS methods on challenging OoD datasets. For NAS-OoD, the optimizer for network parameters is SGD with a learning rate of 0.025. The optimizer for architecture parameters and data generator is Adam. The total number of layers is 20 and the number of initial channels is 36.
We conduct hyper-parameter optimization (HPO) for all baseline methods and compare our NAS-OoD with their performance under the best hyper-parameters. The learning rate and batch size are fixed during HPO. 
For NICO, the experiment setting for all methods follows the protocol of NICO. The network for all baseline methods is ResNet-18. The number of training epochs for all methods is 300. The batch size is 128. We use Adam optimizer with a learning rate of 0.001. Hyper-parameters are selected by HPO with 5 trials. The listed results are the testing accuracy at the time of the highest validation accuracy. The results are averaged over 3 runs with the best set of hyper-parameters.

For PACS, the network for all baseline methods is ResNet-18 initialized with pre-trained weights on ImageNet. As for model selection, we use the leave-one-domain-out method to calculate the testing accuracy for each target domain 
and the average accuracy of these four environments is reported. The number of training epochs is 100. The batch size is 64. We use Adam optimizer with a learning rate of 0.001. Hyper-parameters are selected by HPO with 5 trials. The listed results are the testing accuracy at the time of the highest validation accuracy, averaged over 10 runs with the best set of hyper-parameters.

For the fingerprint dataset, we use ResNet-18 as the backbone network for baseline methods. The number of training epochs is 20. The batch size is 96. Hyper-parameters are selected by HPO with 5 trials. The listed results are the testing accuracy at the time of the highest validation accuracy, averaged over 3 runs with the best set of hyper-parameters.

\section{More Visualization Results of Discovered\\ Architecture Cells}
\begin{figure*}[h]
    \centering
    \subfigure[Normal cell (art painting)]{\includegraphics[height = 3cm]{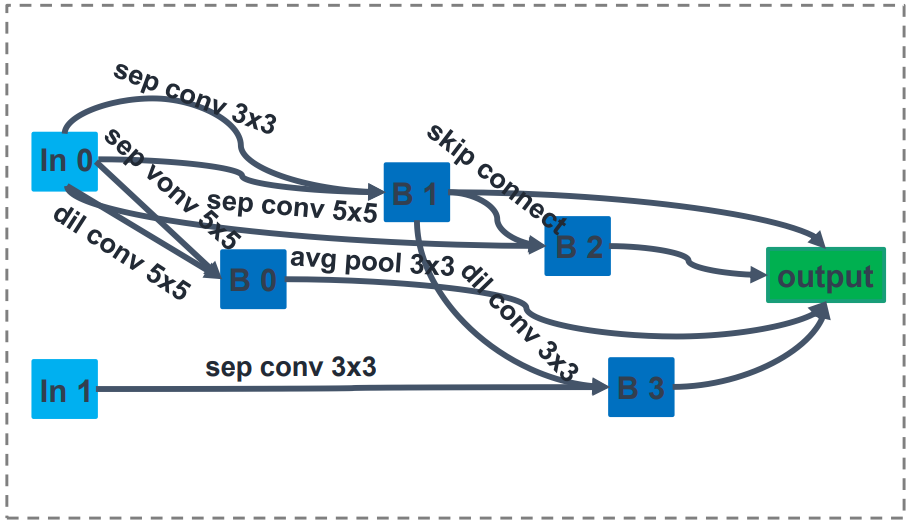}}
    \hspace{1mm}
    \subfigure[Reduce cell (art painting)]{\includegraphics[height = 3cm]{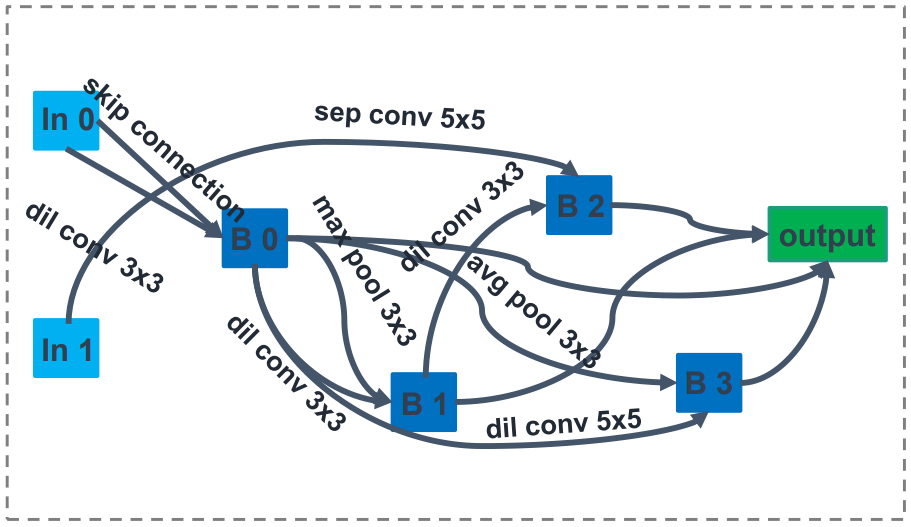}} 
    \hspace{1mm}
    \subfigure[Normal cell (cartoon)]{\includegraphics[height = 3cm]{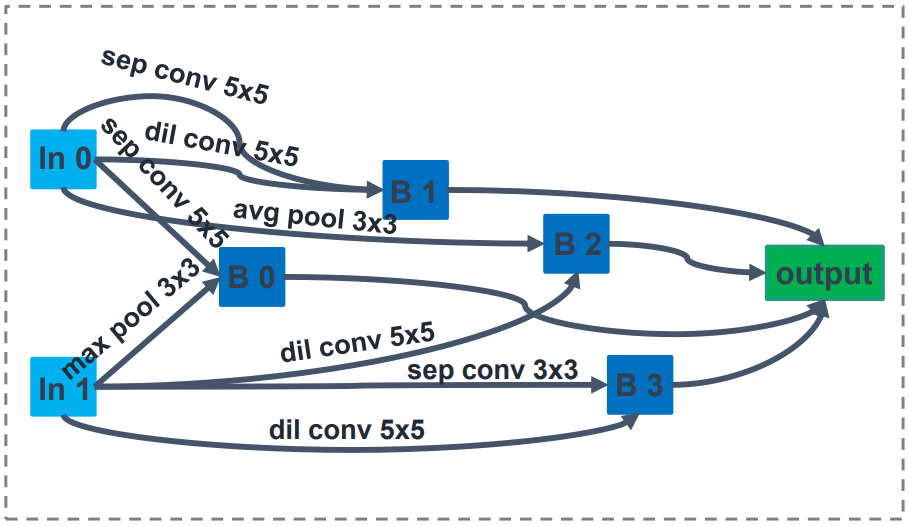}}\\
    \subfigure[Reduce cell (cartoon)]{\includegraphics[height = 3cm]{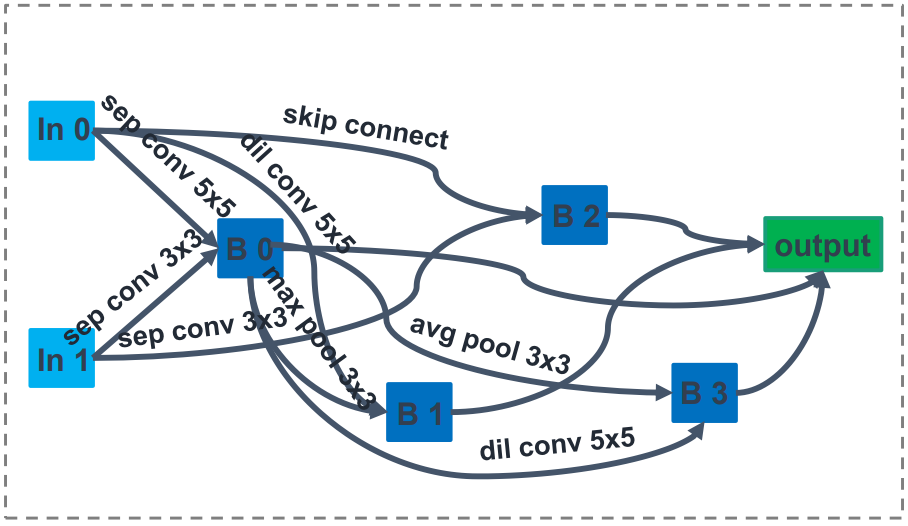}}
    \hspace{1mm}
    \subfigure[Normal cell (sketch)]{\includegraphics[height = 3cm]{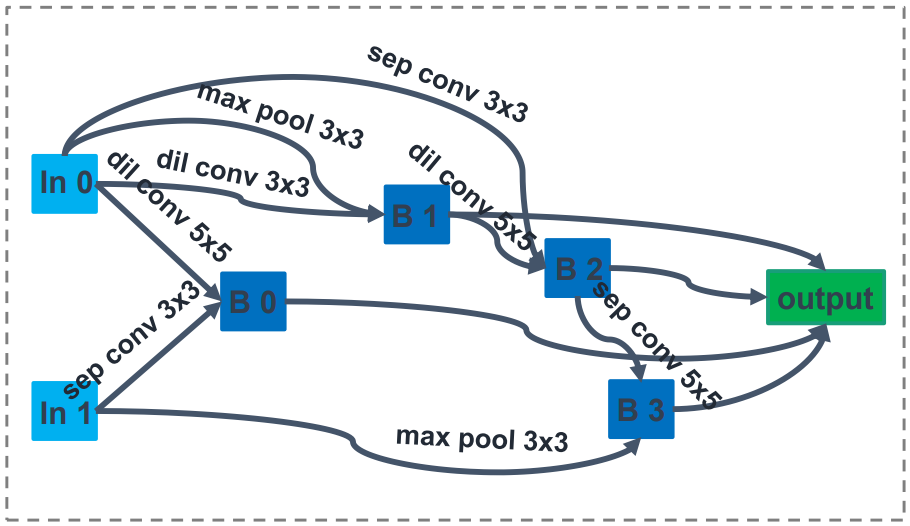}} 
    \hspace{1mm}
    \subfigure[Reduce cell (sketch)]{\includegraphics[height = 3cm]{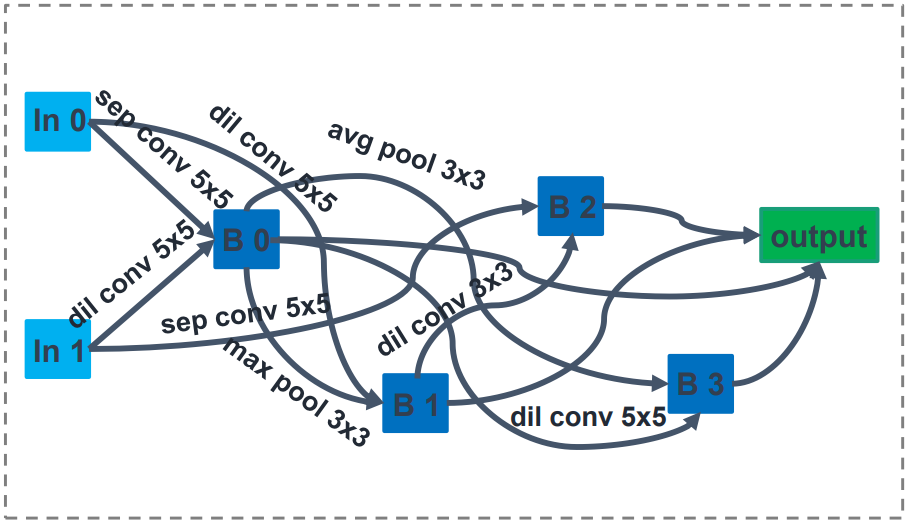}}\\
    \subfigure[Normal cell (photo)]{\includegraphics[height = 3cm]{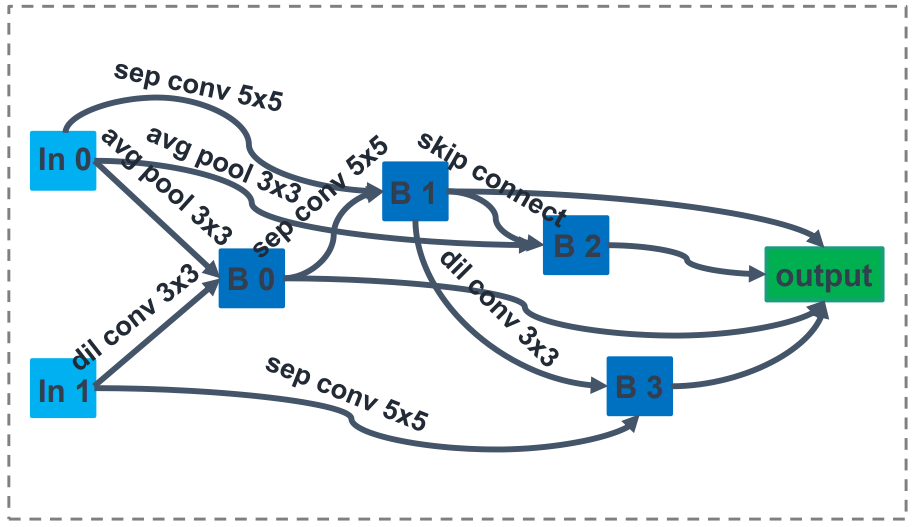}}
    \hspace{1mm}
    \subfigure[Reduce cell (photo)]{\includegraphics[height = 3cm]{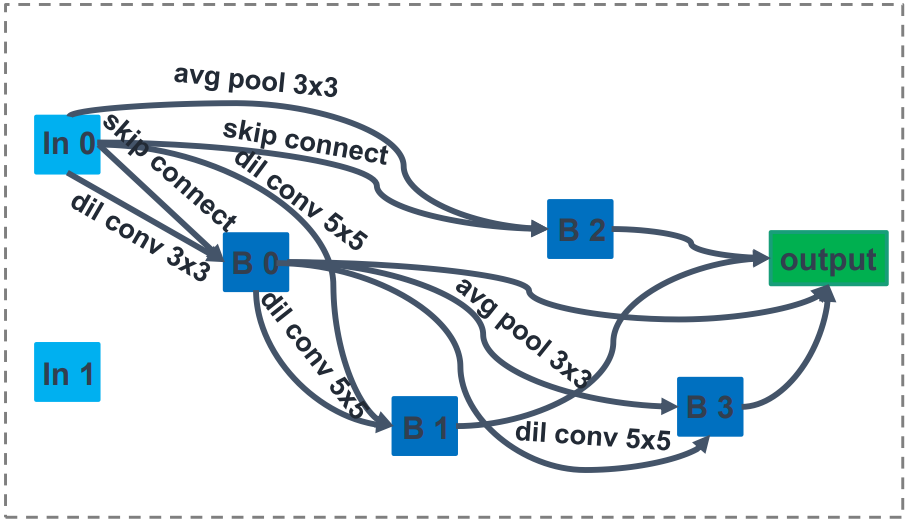}}
    \caption{Typical examples of searched architectures on PACS dataset.}
    \label{fig:pacscell}
\end{figure*}

\begin{figure*}[h]
    \centering
    \subfigure[Normal cell (domain 1)]{\includegraphics[height = 3cm]{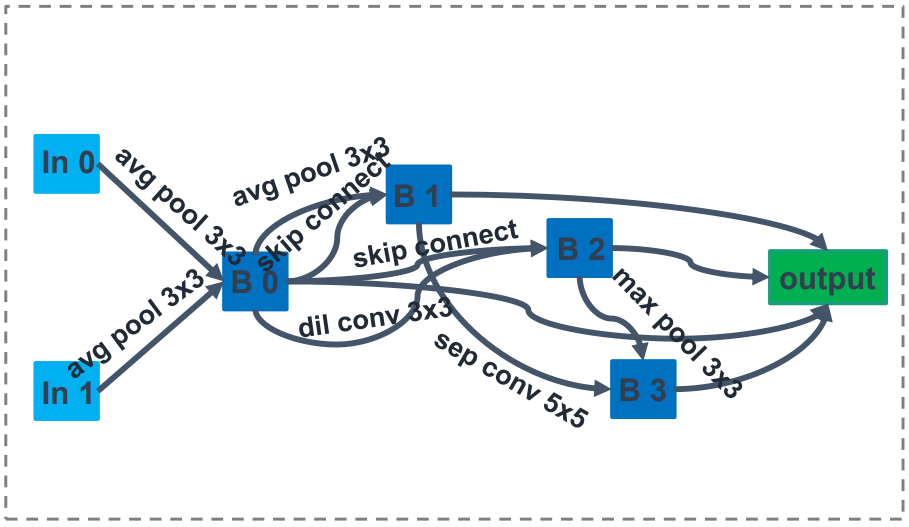}}
    \hspace{1mm}
    \subfigure[Reduce cell (domain 1)]{\includegraphics[height = 3cm]{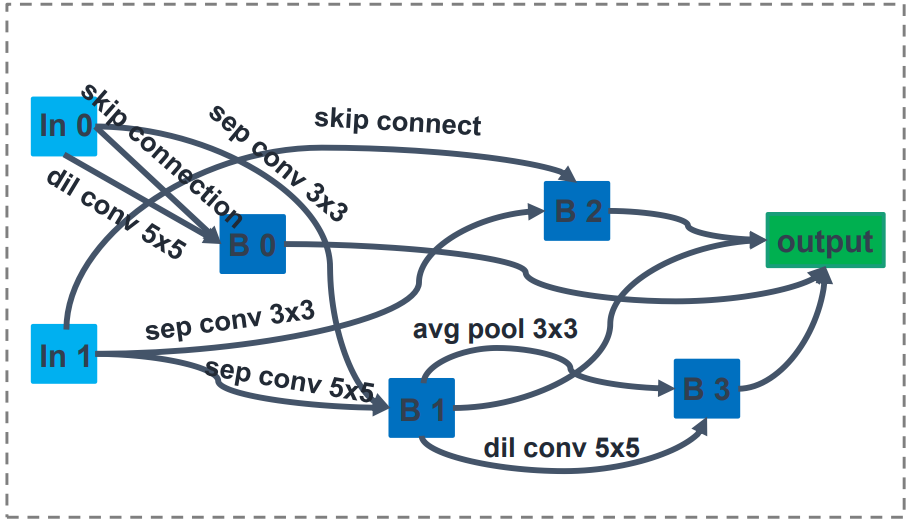}}
    \hspace{1mm}
    \subfigure[Normal cell (domain 2)]{\includegraphics[height = 3cm]{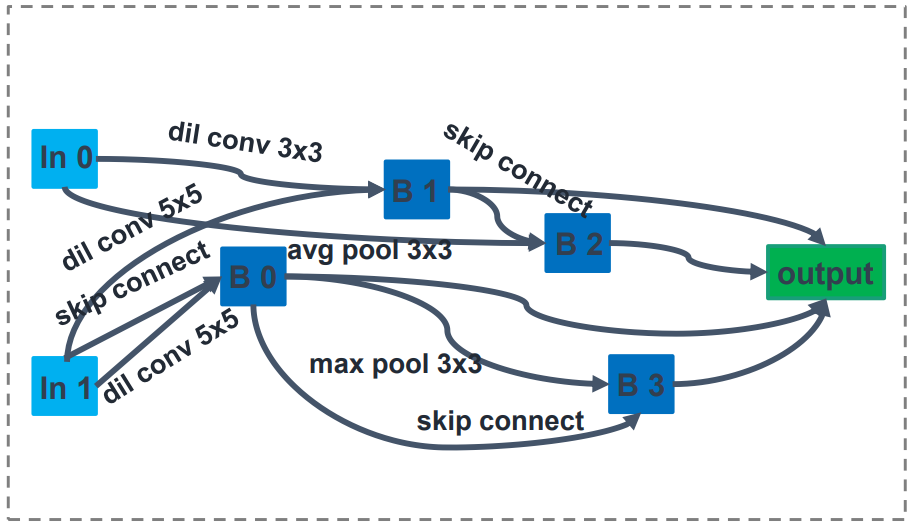}}\\
    \subfigure[Reduce cell (domain 2)]{\includegraphics[height = 3cm]{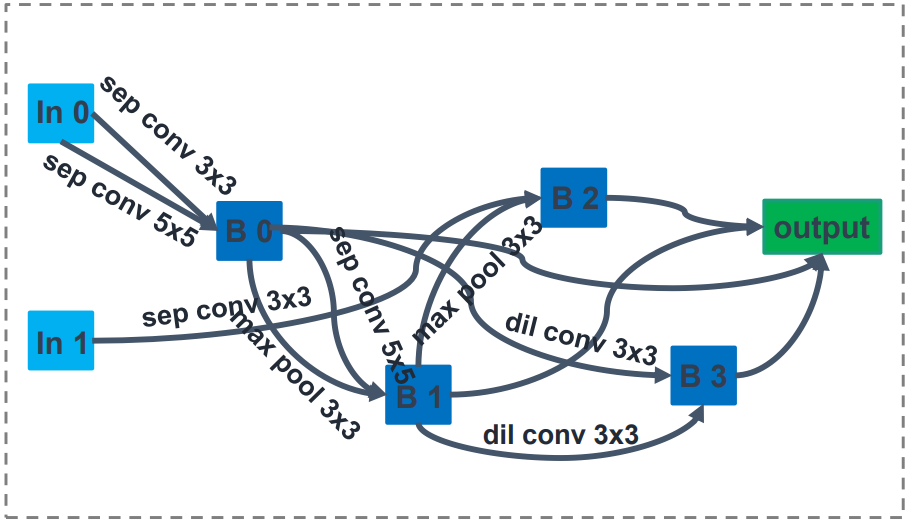}}
    \hspace{1mm}
    \subfigure[Normal cell (domain 3)]{\includegraphics[height = 3cm]{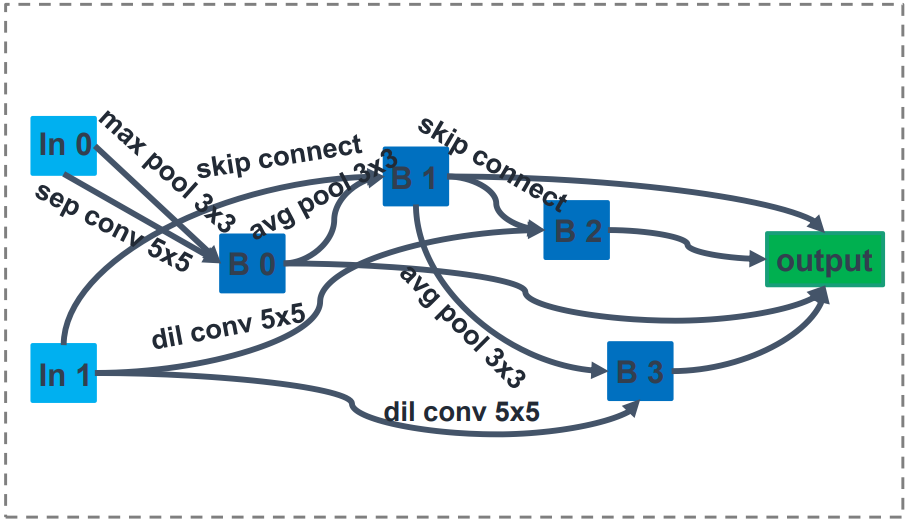}}
    \hspace{1mm}
    \subfigure[Reduce cell (domain 3)]{\includegraphics[height = 3cm]{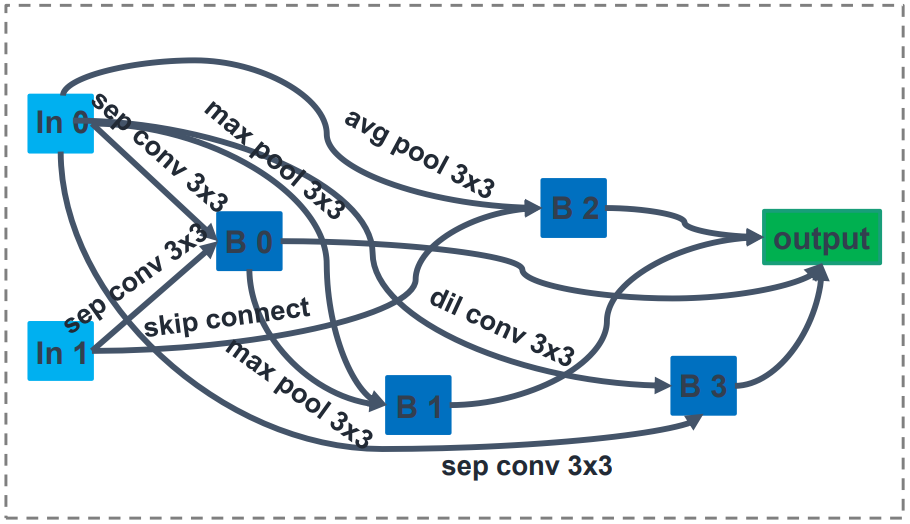}}
    \caption{Typical examples of searched architectures on OoD-FP dataset.}
    \label{fig:fingercell}
\end{figure*}

We also provide more detailed structures of the best cells discovered on different datasets such as PACS (Figure~\ref{fig:pacscell}) and OoD-FP (Figure~\ref{fig:fingercell}) using NAS-OoD.

\end{spacing}

\end{document}